\documentclass{article}

\usepackage{microtype}
\usepackage{graphicx}
\usepackage{subfigure}
\usepackage{booktabs} 

\usepackage{hyperref}

\usepackage[accepted]{icml2019}

\icmltitlerunning{}

\begin{document}

\twocolumn[
\icmltitle{Interactive Naming for Explaining Deep Neural Networks: \\ A Formative Study}




\begin{icmlauthorlist}
\icmlauthor{Mandana Hamidi-Haines}{osu}
\icmlauthor{Zhongang Qi}{osu}
\icmlauthor{Alan Fern}{osu}
\icmlauthor{Fuxin Li}{osu}
\icmlauthor{Prasad Tadepalli}{osu}
\end{icmlauthorlist}

\icmlaffiliation{osu}{School of Electrical Engineering and Computer Science, Oregon State University, Corvallis, Oregon, USA}
\icmlcorrespondingauthor{Mandana Hamidi-Haines}{hamidim@oregonstate.edu}


\vskip 0.3in
]



\printAffiliationsAndNotice{}  

\begin{abstract}
We consider the problem of explaining the decisions of deep neural networks for image recognition in terms of human-recognizable visual concepts. In particular, given a test set of images, we aim to explain each classification in terms of a small number of image regions, or activation maps, which have been associated with semantic concepts by a human annotator. 
%
This allows for generating summary views of the typical reasons for classifications, which can help build trust in a classifier and/or identify example types for which the classifier may not be trusted. 
%
%
For this purpose, we developed a user interface for ``interactive naming,'' which allows a human annotator to manually cluster significant activation maps in a test set into meaningful groups called ``visual concepts''. 
%
The main contribution of this paper is a systematic study of the 
visual concepts produced by five human annotators using the interactive naming interface. In particular, we consider the adequacy of the concepts for explaining the classification of test-set images, correspondence of the concepts to activations of individual neurons, and the inter-annotator agreement of visual concepts. 
%
%
%
We find that a large fraction of the activation maps have recognizable visual concepts,
and that there is significant agreement between the different annotators about their denotations. 
Our work is an exploratory study of the interplay between machine learning and human recognition mediated by visualizations of the results of learning.


%
\end{abstract}


\section{Introduction}

Deep neural networks (DNNs) are powerful learning models that achieve excellent performance on many problems ranging from object recognition to machine translation. 
However, the potential utility of DNNs is limited by the lack of human interpretability of 
their decisions, which can lead to a lack of trust. The goal of this paper is to study an approach, called \emph{interactive naming}, for improving our understanding of the decision-making process of DNNs. In particular, this approach allows a human annotator to visualize and organize activation maps of critical neurons into meaningful visual concepts, which can then be used to explain decisions made over the test data. 

Interpreting the roles and functions of individual and groups of neurons in DNNs in order to explain their decision making has been a long-standing problem in AI \citep{Hinton:PDP}. Much recent work on interpretibility has focused on visualizing activation maps which highlight parts of the input that are most important to the final decision of the DNN or the output of an individual neuron. These include backpropagation-based methods, e.g., \citep{MatDeconv,SimonyanVZ13,JTguided2015,Avanti16,integrad17,LRP15,Avanti16,zhang16excitationBP,Gradcam17}, which compute the activations for all input features in a single forward and backward pass through the network; and perturbation-based methods, e.g., \citep{MatDeconv,Dabkowski2017,FongMask17}, which perturb parts of the input to see which ones are most important to preserve the final decision. While activation maps can be a useful tool for analyzing DNNs, they do not necessarily identify the human recognizable concepts embedded in the network, nor do they provide an understanding for how those concepts combine to form the final decisions. 

In this work, we make progress toward this goal by building an interface for interactive naming, and conducting a formative study on a set of non-trivial image classification tasks. In particular, our approach is based on the idea that the final decision of a DNN is dominated by the most highly-weighted neuron activations (the \emph{significant activations}) in the penultimate network layer. Explanations of the decisions can thus be formed by 1) identifying the significant activations for each decision,  and 2) attaching meaningful concepts to the significant activations. Since DNNs typically have thousands of units in the penultimate layer, (1) can result in an overwhelming number of activations. To address this issue we draw on recent work that augments the original DNN with a learned \textit{explanation Neural Network} (\textit{xNN}), which mimics the predictions of the DNN using a much smaller penultimate layer of ``X-features". Since the xNN is effectively equivalent to the original DNN, we can use it to make predictions on test instances with no loss in accuracy, but with a dramatic reduction in the number of significant activations to be considered for explanations. 

To deal with (2), our interface displays the (significant) activation maps of X-features for decisions made on a test set and allows a human annotator to cluster the activations into meaningful groups called ``visual concepts.'' 
Even though there are a small number of significant activations that sufficiently explain the final decisions, 
there may not be a one-to-one correspondence between them and human-recognizable visual concepts. 
Indeed, unlike in the standard supervised learning setting, where the number of classes/concepts is typically fixed beforehand, the number of visual concepts covered by the set of all significant activations is unknown. To make matters more interesting, the set of visual concepts might be different for different annotators. Finally, 
the annotators may not be able to label a map in isolation, and might need to 
see multiple images and find similarities and differences before labeling them. Indeed, this 
last problem has been studied under the name of ``structured labeling,'' 
in the context of active learning and provides an inspiration for our work \citep{Kulesza2014}.

Drawing from the lessons of the above work, our interface provides maximum flexibility to the human annotators by presenting them with the activation maps of all X-features of all test images that belong to each category. Unlike the previous work 
on supervised and active learning which seek labels from a fixed label set, 
the annotators are asked to cluster the maps in a way that makes most sense to them and give them meaningful names. The subjects can 
create new clusters, move images across clusters, 
and merge clusters. They are also allowed to leave some maps not clustered, and discard others that they 
consider noisy. 

The result of interactive naming is a set of explanations of test set predictions in terms of visual concepts. This enables summarizing the types of predictions that are made to gain confidence in the predictor and/or identify potential flaws in the predictor.
Importantly, this type of summary is dependent on the human annotator, which raises interesting questions about differences in explanations that might result from different annotators. Specifically, we seek answers to the following research questions (RQs) through our study:\\



\textbf{RQ1 (Coverage of Interactive Naming)}: What fraction of the examples are explainable using human recognizable visual concepts? If a significant fraction of the examples are not explainable via visual concepts, it might mean that the X-features 
are not properly aligned with human concepts and will have to be retrained from human data. 

\textbf{RQ2 (Correspondence Between X-features and Visual Concepts)}: Is there strong correspondence between visual concepts and individual X-features? In the absence of strong correspondence, we might  
explain the final decision of the network directly in terms of visual concepts rather than in terms of the X-features which are in turn interpreted by visual concepts. 


\textbf{RQ3 (Inter-annotator Agreement)}: 
How much overlap exists between the annotated sets of activations between different subjects? How 
much do the clusters of different subjects overlap?
Existence of significant overlaps might suggest that we can move toward building a standardized ontology of visual concepts for explanations. Lack of significant agreement might 
mean that we will have to personalize 
explanations to different annotators.

We explore the above questions through 
empirical experiments and annotator 
studies based on data from 5 annotators on a bird species classification dataset with 12 different species of birds \citep{WahCUB_200_2011}. 
The annotators are asked to name the activation maps of different bird images with  
human-recognizable visual concepts with no attempt to recognize their species. The studies reveal that a significant fraction of the images are human recognizable with some individual differences among different annotators.

\section{Related Work}

A lot of the current work on explaining the decisions of deep neural networks is focused on visualizing the activation maps of the neurons computed from various methods \citep{MatDeconv,SimonyanVZ13,JTguided2015,Avanti16,integrad17,LRP15,Avanti16,zhang16excitationBP,Gradcam17}. 
Two recently proposed explanation techniques, PatternNet and PatternAttribution,  produce better visualizations 
for linear and nonlinear models than the earlier Gradient, DeConvNet and GuidedBackprop methods \citep{Kindermans2017_IUI}.
In contrast, a new perturbation-based approach identifies relevant patches in the input image by sampling the image through perturbations, identifies the important neurons, and then passes them through a deconvolutional neural network \citep{Lengerich2017_IUI}.
Unfortunately, most of these activation map-based methods only provide qualitative results as they do not try to associate activations with human-recognizable concepts.

Another class of approaches generates 
visual explanations using natural language justifications, e.g. see  \citep{Lisa2016_IUI}. 
However, sentence-based explanation approaches require an extra corpus of sentence descriptions in the training process. A different line of work, e.g. \citep{ribeiro2016_IUI}, attempts to extract rule-based explanations of DNN decisions. This is done via analyzing the impact of perturbations to the input and extracting a rule that ``locally explains" the network decision for a given input. \citep{Lundberg2017_IUI}
identifies the class of feature importance methods and shows that there is a unique solution in this class that satisfies some desirable properties.
One challenge for using these approaches is that they require predefined concepts to be used for explanations.

There has also been some recent work which analyzes the alignment between individual hidden neurons and a set of semantic concepts \citep{netdissect2017}. While this provides additional insight into the semantics of neurons, it requires large sets of data labeled by the semantic concepts and is limited to the semantic concepts in that data. This approach also does not attempt to relate the concepts to final DNN predictions. 
A more recent approach disentangles the evidence encoded in the activation feature vector, and quantifies the contribution of each piece of evidence to the final prediction \citep{BoleiZhou2018_IUI}. However, similar to previous work, it is also based on supervised learning of predefined concepts and cannot identify concepts which are not learned beforehand. 

XNN\citep{QiXNN}, a recent explanation approach, attempts to do away with predefined concepts. It learns a nonlinear dimensionality reduction from the deep learning feature space to a low-dimensional space and tries to mimic the deep learning prediction using this low-dimensional space. It is very faithful to the deep learning classifier it is attempting to explain, however, sometimes the learned concepts cannot be recognized by human. In this paper we build on XNN and attempt to obtain human-understandable visual concepts via interaction with the explanation.

Importantly, none of the current approaches support human interaction in recognizing, clustering, and naming the concepts implicitly employed by the neural network in making its decisions.
While some methods do employ human recognizable concepts, they are learned by the system offline from a large amount of labeled data that may or may not be relevant to the task at hand.  

\begin{figure*}[h!t]
\begin{center}
     \includegraphics[height=8.5cm]{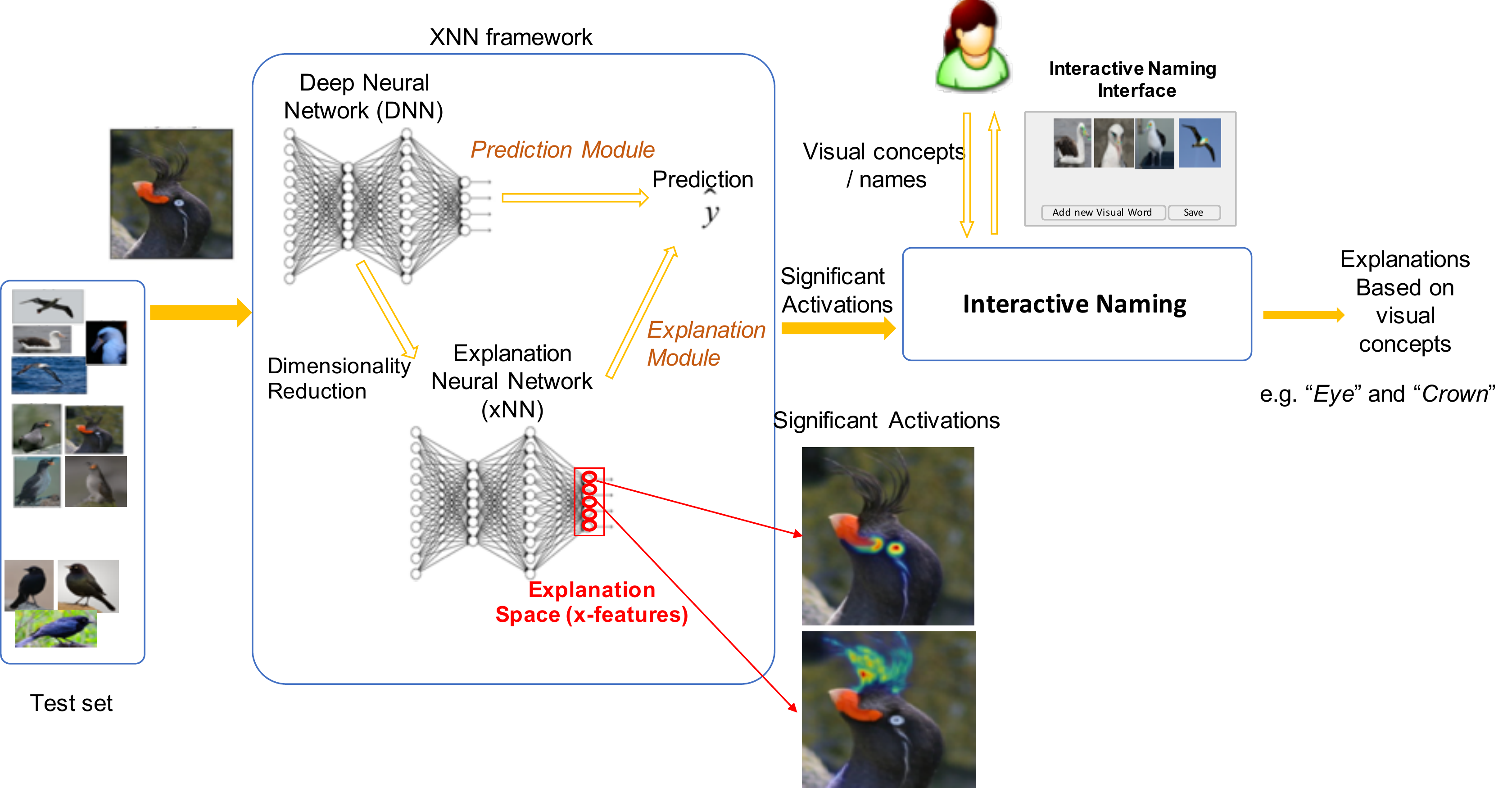}
\caption{\small Interactive Naming Framework. Conceptually, the explanation module is a dimensionality reduction mechanism so that the original deep learning prediction $\hat{{y}}$ can be reproduced from this low-dimensional space. An explanation module can be attached to any layer in the prediction deep network (DNN). The output of the DNN can be faithfully recovered from this low-dimensional explanation space, which represents high-level features that are interpretable to humans. }
\label{fig:InteractiveNaming}
\end{center}
\end{figure*}

\section{Interactive Naming for Test Set Explanations}

We first give an overview of the overall approach and then describe each component of the system. 

\subsection{Overview} Our overall goal is to develop tools to help understand the decisions of deep neural networks (DNNs) that are trained for image recognition via supervised learning. In particular, we aim to generate meaningful explanations for decisions made over a representative set of test images. This can provide insight into the strengths and weaknesses of the learned DNN that may not be apparent by just observing test set accuracy. For example, one might hope to discover situations where the DNN is making the right decision, but for the wrong reason, which would identify potential future failure modes. 

Figure~\ref{fig:InteractiveNaming} give an overview of our \emph{interactive naming} approach for producing test set explanations. At a high-level, each DNN decision for a test image is dominated by a set of the most \emph{significant activations} of neurons in the penultimate layer. Thus, attaching meaningful concepts to those activations is one way to explain decisions. However, typical DNNs use very large penultimate layers, which makes training easier, but can result in less compact explanations due to the large numbers of significant activations. For this reason we attach an \emph{explanation neural network (xNN)} to the penultimate layer of the DNN, which is trained to reproduce the decisions of the DNN, but dramatically reduces the number of significant activations. The xNN is thereafter used to make decisions and explanations can be formed in terms of its much smaller number of significant activations. 

In order to attach meaning to the significant xNN activations we developed an interactive naming interface which displays visualizations of the significant activations to a human annotator. The annotator is then able to cluster the activations into meaningful groups, called visual concepts, and attach linguistic labels to the groups if desired. Given a test instance, we can then form an explanation by producing the significant xNN activations and displaying the group identities/names of those activations. Qualitatively different decisions will tend to have different explanations. A key functionality of the system is to allow for the investigation into the different qualitative decision types over the test set.  

The rest of this section explains the above steps in more detail. 


\subsection{Explanation Neural Networks (xNNs)}


An xNN \citep{QiXNN} is an additional network module, which can be attached to any intermediate layer of an original DNN, which typically have thousands of neurons. The xNN learns a 
lower dimensional embedding for the DNN layer, resulting in a vector of \emph{X-features}, and then linearly maps the X-features to the output $\hat{y}$ in order to mimic the output $y$ of the original DNN model. In this work we focus on xNN modules that are attached to the first fully-connected layer of the orignal DNN, which reduces thousands of features in the penultimate later to a handful of X-features. 

In our work, we will be applying xNNs to DNNs for multi-class classification. To do this, as proposed in \citep{QiXNN} a distinct xNN is trained for each class in a one-against-all manner to predict the numeric score assigned to the class by the DNN. The xNNs can then be used for multi-class prediction by computing the scores produced by each xNN and returning the highest scoring class. 

It is desirable for X-features to have the following 3 properties: 1) \textit{faithfulness}, the DNN predictions can be faithfully approximated from a simple linear transform of the X-features; 2) \textit{sparsity}, a relatively small number of X-features are active per image, 
and 3) \textit{orthogonality}, the X-features are as independent from each other as possible. Training the xNN thus involves optimizing an objective function via backpropagation that includes terms for each of these properties \citep{QiXNN}. Details are beyond the scope of this paper. 


\begin{figure*}[h!]
\centering
    \begin{tabular}{c}
       \includegraphics[height=5cm]{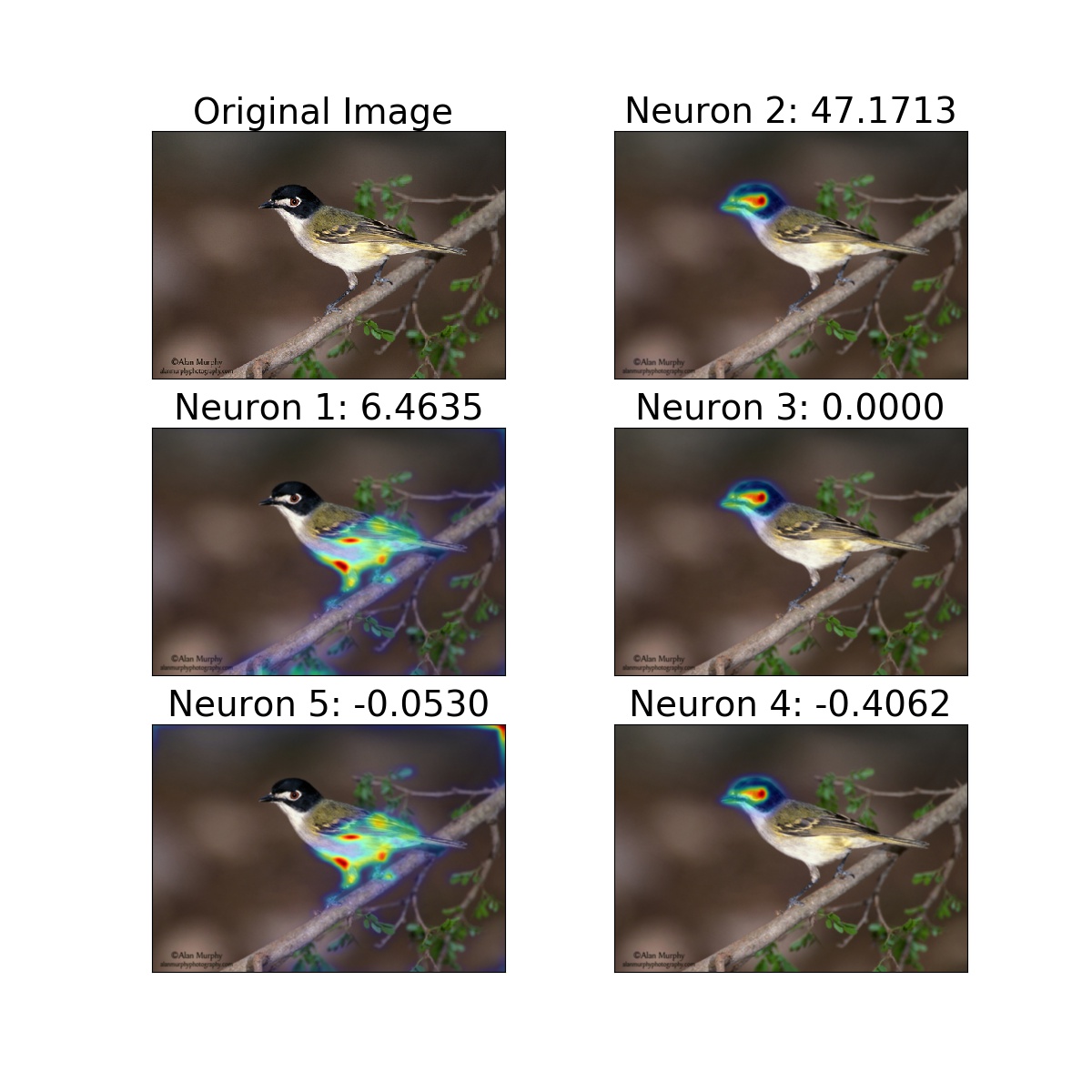}
        \includegraphics[height=8cm]{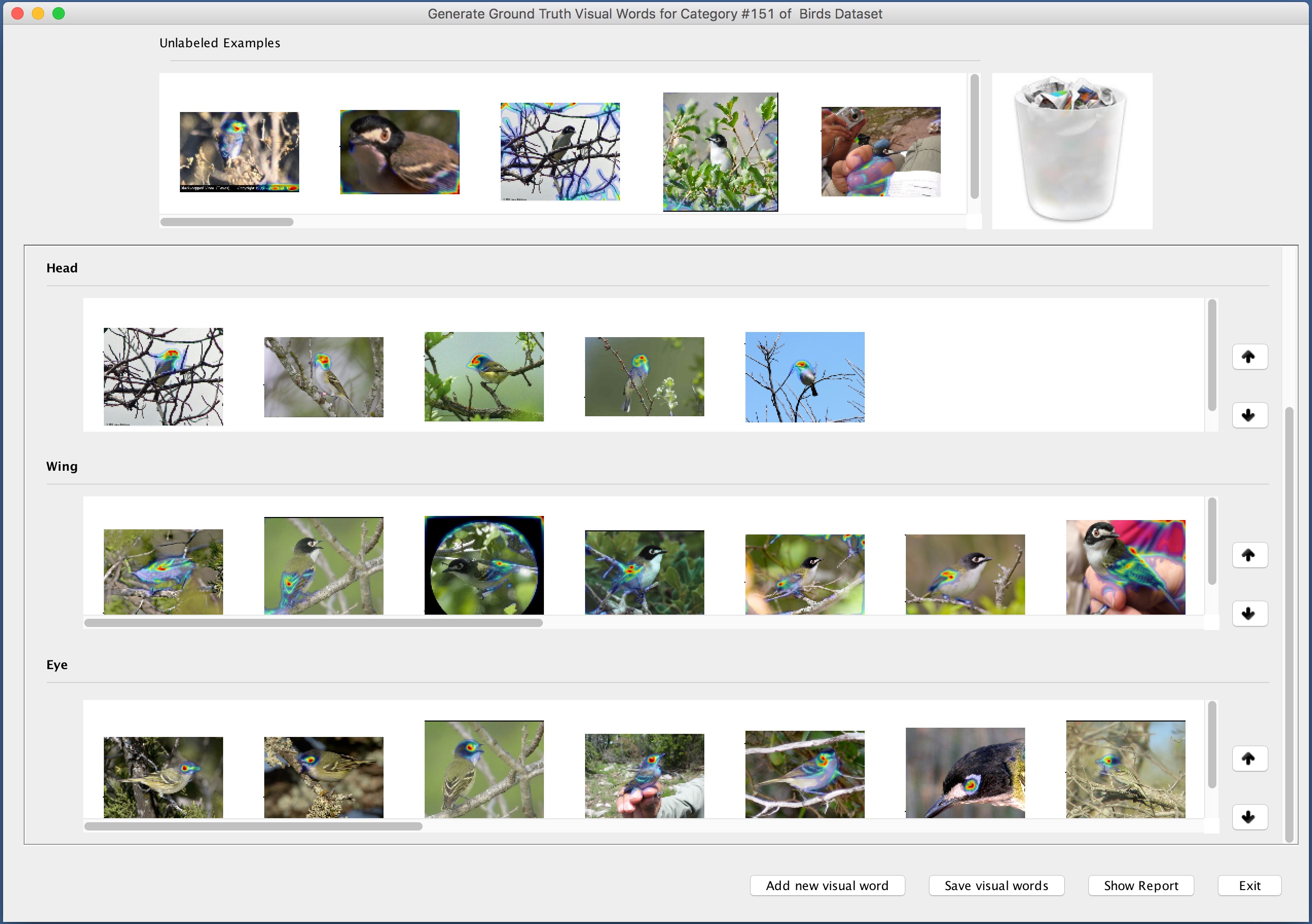}
    \end{tabular}
   \begin{tabular}{c}
    \end{tabular}
\caption{\small (Left) Examples of visualization of x-feature activations.  (Right) Annotation Interface: Our approach allows annotators to explore feature activations and group them into different meaningful textual / visual concepts.}
\label{fig:attnetionMapsandUI}
\end{figure*}


\subsection{Explanations via Interactive Naming}

Given a test image and a class $c$, we can use the xNN for $c$ to produce a class score. This score is a linear combination $\sum_i w_i\cdot x_i$ of the X-features $x_i$ and their associated weights. The positive terms (i.e. X-features with positive weights) in the linear combination sum to provide a positive score that can be viewed as providing positive evidence for $c$. Typically only a subset of the positive terms are significant. Thus, we define the \emph{significant X-features} for the image to be minimum subset of X-features that account for at least 90\% of the positive score. The significant X-features can be viewed as a type of explanation of why the image might be assigned to class $c$. However the significant X-features do not have associated semantics, so the explanation is not very useful for human consumption. 

To assign semantics to explanations, we can first produce an \emph{activation map} for each significant X-feature in an image for the class under consideration, which identifies the ``salient" image region that is responsible for the X-feature activation. In this work, we use the ExcitationBP algorithm for computing activation maps \citep{zhang16excitationBP}. We call these maps the \emph{significant activation maps} or simply the \emph{significant activations}. While one can gain insight into a prediction by simply viewing the significant activations, it is difficult to obtain a general understanding of the core semantic concepts and combinations of those concepts used for predictions across an entire test set, which is our goal. 

Our interactive naming interface is designed to attach semantics to all of the significant activations across a test set. In particular, the interface allows a human annotator to cluster, or group, the significant activations in a test set of images, where each group is intended to represent a semantically meaningful \emph{visual concept} to the annotator. Not all significant activations need to be assigned to a group, which allows for noisy or confusing activations to be handled. Activations that are assigned to a visual concept are considered to the \emph{named}, while other activations are considered to be \emph{unnamed}. The complete set of named activations resulting from interactive naming is called a \emph{naming} of the test set. 

Given a naming of a test set, we can now generate an \emph{explanation} for each test image by generating the significant activations of the image and outputting the visual concept names for those activations. Thus, an explanation is just a set of names. If a significant activation is unnamed, then the explanation includes ``other" for the name of the activation. 

\subsection{Interactive Naming Interface} 

One of the key aspects of interactive naming is that the set of visual concepts is not known beforehand and varies from person to person. Moreover, the visual concepts in an image are not immediately apparent until the 
annotator sees multiple images. In previous work, it has been shown that 
human labelers are more efficient when they are presented with multiple instances at once and are allowed to 
choose the ones they want to label \citep{Kulesza2014}. In another  
real-world comparative study of pairwise and setwise comparison, the authors demonstrated that not only is setwise comparison more efficient, but also elicits more consistent labels \citep{Sarkar_CHI2016}.

Following the previous work, 
we designed a flexible user interface(Fig~\ref{fig:attnetionMapsandUI} (Right)) to group the significant activations into different visual concepts and give them textual labels/names. 
The set of X-feature activations 
is shown to the annotator in the ``Unlabeled Examples'' section of the interface. 
The annotator can freely cluster  
X-feature activations into visual concepts and give them names. 
The interface allows the annotators to compare all instances,
and create visual concepts as and when they are confident. If the 
annotator is not comfortable with 
grouping or labeling some activations, 
they can leave them in the unlabeled section. 



\begin {table*}[h!t]
\centering
\begin{tabular}{|l|c c c c c c c c c c c c|} 
\hline 
Index of category &\textit{a} & \textit{b} &\textit{c}& \textit{d}& \textit{e} & \textit{f}& \textit{g}& \textit{h}& \textit{i} & \textit{j} & \textit{k} & \textit{l} \\
\hline
\hline
Number of images  & 60 &  44    &  59 &    60     & 60  &   57    & 60        & 60     & 60   &  60 &   60  & 51\\
\hline
Total significant activations & 118 & 108 & 158    & 120 & 73 & 167 & 138 & 125 & 170 & 124 & 120& 81 \\
\hline
Average of significant activations & 1.97 & 2.45 & 2.68 & 2 & 1.22 & 2.93 & 2.3 & 2.08 & 2.83 & 2.07 & 2 &  1.49 \\
\hline
dNN  accuracy (\%) &86.7 & 93.2 & 67.8 & 95.0 & 88.3 & 94.7 &88.3  &75.0  &65.0  &75.0  & 91.67 & 96.1  \\\hline
xNN  accuracy (\%) &86.7 & 93.2 & 67.8 & 95.0 & 88.3 & 96.5 &88.3 &75.0  &65.0  &75.0  & 91.67 & 96.1  \\\hline
xNN  RMSE  &0.23 & 0.23 & 0.51 & 0.41 & 0.35 & 0.30 &0.50  &0.20  &0.21  &0.34  & 0.45 & 0.33  \\
\hline
\end{tabular}
\caption { The relevant X-feature activations of 12 bird categories: \textit{(a) Laysan Albatross, (b)Crested Auklet, (c) Brewer Blackbird, (d) Red-winged Blackbird, (e) Northern Fulmar, (f) Green Jay, (g) Mallard, (h) Black Tern, (i) Common Tern, (j) Elegant Tern, (k) Green-tailed Towhee}, and \textit{(l) Black-capped Vireo}. The table shows the number of images for which each feature makes a significant positive contribution.} 
\label{table:dataset} 
\vspace{-1em}
\end {table*}


\subsection{Data Preparation}
All our 
experiments were conducted on the  
Caltech-UCSD Birds-200-2011 dataset \citep{WahCUB_200_2011}. 
There are 12 categories of birds 
labeled a-l in Table~\ref{table:dataset}. The first row of the table shows the number of images in each category.  
We begin with a convolutional deep neural network (DNN) trained on the available multi-class data. The DNN outputs $p(c_i|I)$ for each given image $I$ and category $c_i \in 1, \ldots, C$. The penultimate layer of the DNN can be considered as scoring functions for each category $s(c_i|I)$, where a softmax unit $p(c_i|I) = \frac{s(c_i|I)}{\sum_{i=1}^C s(c_i|I)}$ serves as the final layer of the DNN that computes the class-conditional probability from the scores. xNN is trained starting from the first fully-connected layer in the DNN for each class, aiming at being faithful to the scoring functions $s(c_i|I)$ for each category. 

As reported in 
\citep{QiXNN}, this approach reduces the dimensionality 
from 4,096 features in the DNN to 5 X-features in the xNN without significant loss of accuracy. 
The last two rows of Table~\ref{table:dataset} shows the multi-class classification accuracy of the xNN on the original $200$ categories after replacing the DNN $s(c_i|I)$ with the one generated by xNN for each respective category, as well as the original DNN accuracy on those categories. It can be seen that xNN has almost identical accuracies as the DNN, even performing better than DNN in one case. xNN also approximates exact scores of DNN well, which is shown in the table in terms of root mean squared error (RMSE). One can see that xNN has an RMSE between $0.2 - 0.5$ while the range of the scoring function is usually $0 - 50$.

To enable human annotators to recognize and name the X-features, 
we need to visualize the activation map of each X-feature on input images. This is done using 
ExcitationBP\citep{zhang16excitationBP}, a gradient-based visualization approach. We applied ExcitationBP on each 
X-feature respectively and generated the activations for this X-feature.
Fig~\ref{fig:attnetionMapsandUI} (left) shows an example of a bird's original image followed by its 
5 X-feature activations. The X-feature activations superimposed on the original image 
are presented as ``activation maps'' for visualization. 

Since the X-features with negative weights do not provide positive evidence to the class at hand,
their activation maps are not used for annotation. We further filter the activation maps to only those maps that contribute to 90\% of the total positive weight for the final decision. We call these \textit{significant activations}. The second row of 
Table~\ref{table:dataset} shows the total number of significant activations in each category. The third shows the average number of significant activations per image. 

\section{Human Subject Study}
We had the activation maps of the different images annotated by 5 different 
subjects using the annotation interface. The activation maps were separated 
by the class, but not by the X-feature. The annotators were instructed to not 
introduce visual concepts that only applied to one or two images, but were otherwise free to cluster and label as many images as it 
made sense to them.  However, not all subjects followed 
instructions and left some clusters with less than 3 images. In the following analysis, we first cleaned the data by removing a 
small number of clusters with less than 3 images. 


\subsection{RQ1: Coverage of Interactive Naming}

Since the human annotators are not forced to assign visual concepts to, or name all significant activations, some of the activations in the data are unnamed and treated as noise/outliers. Here we are interested in how well the annotations cover the activations and explanations and how this coverage varies across annotators. 

\begin{figure}[h!]
\centering
\includegraphics[width=8cm]{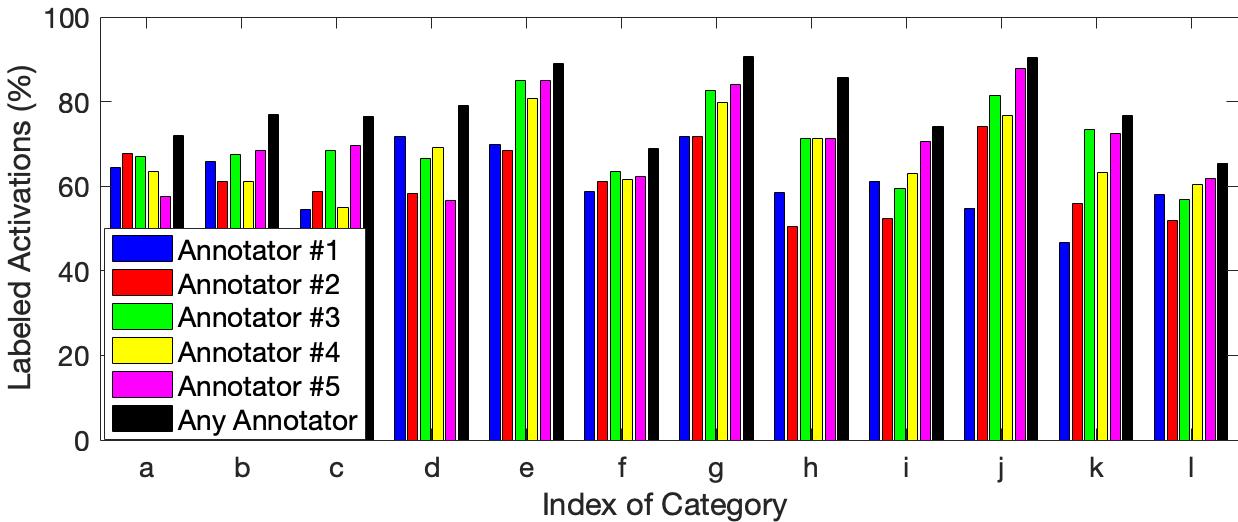}\\
\caption{Fraction of labeled significant activations for each category across annotators.}
\label{fig:FractionOfLabeled_activationMaps}
\vspace{-1em}
\end{figure}

Figure \ref{fig:FractionOfLabeled_activationMaps} shows the fraction of significant activations that are named by each annotator for each bird category. In addition, the last bar for each category, labeled ``Any Annotator", shows the fraction of significant activations that were assigned to a visual concept by at least one annotator. We see that within a particular class, there is relatively small variation among users and that the ``Any Annotator" bar is not much higher than that of the typical individual annotator. This indicates that there is some consistency in the set of activations that users consider to be noise. We also see that for most categories there is a relatively significant amount of activations not labeled by users, approximately ranging from 20\% to 40\%. 

We now consider how well the annotations cover explanations, which gives a better sense of how useful they will be for analyzing explanations. In particular, we consider an explanation for an image to be \emph{completely (partially) covered} by an annotation if all (at least one) of the significant activations for that image are named. 

\begin{figure}[h!]
\centering
\includegraphics[width=7.6cm]{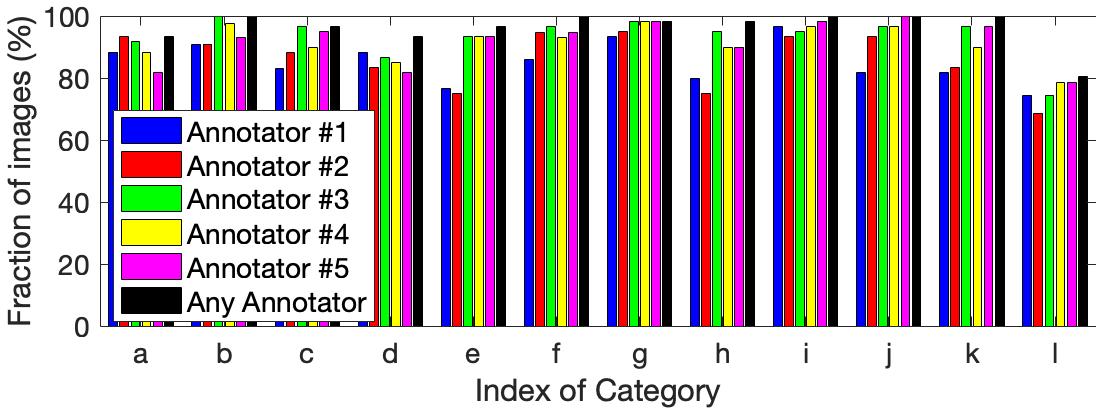}\\
\caption{Partial explanation coverage for each category across annotators. }
\label{fig:partialCoverage}
\vspace{-1em}
\end{figure}

\begin{figure}[h!]
\centering
\includegraphics[width=7.6cm]{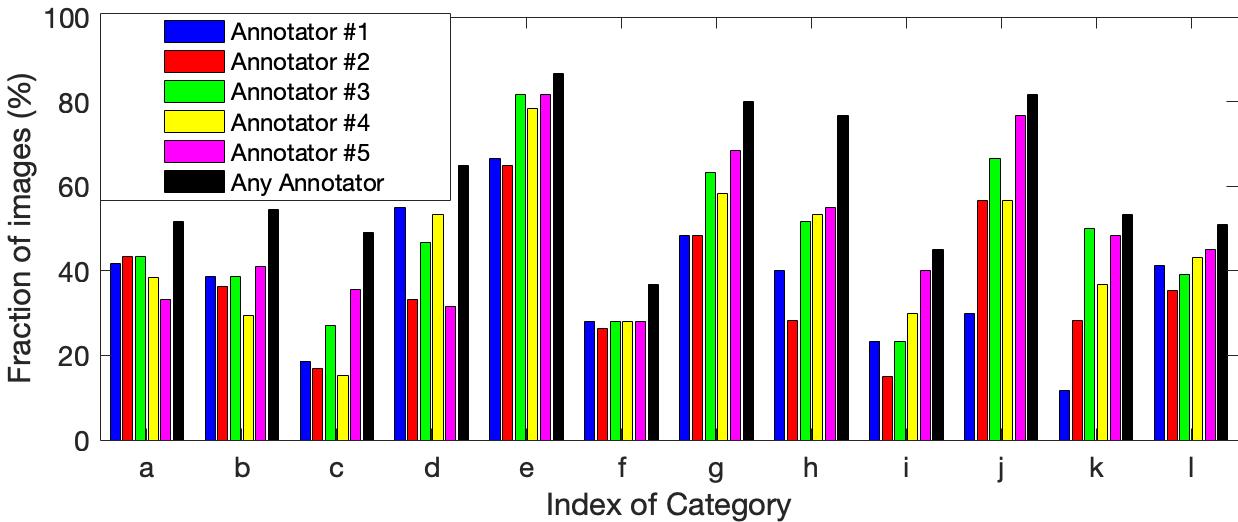}\\
\caption{Complete explanation coverage for each category across annotators.}
\label{fig:completeCoverage}
\vspace{-1em}
\end{figure}

Figures \ref{fig:partialCoverage} and \ref{fig:completeCoverage} show the partial and complete coverage for each annotator and the ``Any Annotator". We see that for most annotators the fraction of explanations that are at least partially covered is quite high. This means that at least partial semantics will be available for explanations on the vast majority of cases. We also see that the ``Any Annotator" bar is similar to the individual annotators, which indicates that the sets of partially covered explanations across annotators is similar. The complete coverage percentages drop substantially, which is not surprising given the results for activation coverage from Figure \ref{fig:FractionOfLabeled_activationMaps}.
Once again the ``Any Annotator'' bar is not significantly different from the rest. 

Overall, we see that a non-trivial fraction of significant annotations are not named by users. This necessarily resulted in less than 50\% complete coverage rates for most users. 
From further analysis, the low rates of 
complete coverage seems to be due to the fact that in most cases the final decision is dominated by a single significant activation. The other significant activations appear to be not strong enough to lead
to easily recognizable concepts and hence are not named by the annotators. Still there are a non-trivial number of examples that are completely covered, which allows for analysis of full explanations for a fraction of the test data. Even this can help build trust and identify potential flaws. On the other hand, most explanations are at least partially covered, which allows for gaining some semantic insight into most decisions. Even partial explanations can be useful in identifying flaws and building trust. These results suggest future work on increasing the coverage by bootstrapping from partial coverage, e.g. by incorporating the annotations into retrained DNNs. 

We performed a qualitative analysis to understand some of the reasons that annotators were not able to assign names to activations. One of the major reasons was when activations were difficult to interpret and appeared to be noise. For example, when activations highlight the edge of the image or fall on background with unclear semantics. Such activations are potential warning indicators about a classifier. Thus, uncovering these examples through interactive naming has value. In other cases, the activation map was interpretable to the annotator, but there were not enough similar activation maps to form a cluster. This case may be resolved by using a larger test set. 

\subsection{RQ2: X-feature-Visual Concept correspondence}
We now consider the question of 
correspondence between the visual concepts and the 
X-features. Since one of the goals of xNNs is to disentangle 
the concepts into independent components, we wanted to see
how well this goal has been achieved. In an ideal case, one 
might hope for a one-to-one correspondence between visual 
concepts and the X-features. However, this is unrealistic 
as our xNN did not have any supervision on the visual concepts. 

We adopt the metric of \emph{purity} to measure the correspondence between visual concepts and X-features \citep{Manning:2008}. The purity of a clustering is defined as the number of examples that belong to the
plurality class of each cluster as a fraction of the total number of examples. Here, we employ purity to measure both the degree to which each visual concept maps to a single X-feature, i.e., visual concept $\rightarrow$ X-feature purity or \emph{CX-purity},  and the degree to which each X-feature maps to a single visual concept, i.e., X-feature $\rightarrow$ concept or \emph{XC-purity}. 

We call the set of clusters of activations 
created by each annotator, {\em a naming}. 
To compute CX-purity of a naming, we first assign each visual 
concept to the X-feature to which a plurality of its
significant activations belong. We call it \emph{the  
majority X-feature} of that concept. CX-purity is the 
number of activations in the naming that belong to the
majority X-feature of their concept as a fraction of all named 
activations. Similarly we define the majority concept of an X-feature  to be the 
concept that is assigned to a plurality of activations of that 
X-feature. XC-purity is the number of activations in the naming that belong to the majority concept assigned to their 
X-feature as a fraction of all named activations. 

As it may be clear, both CX-purity and XC-purity vary from 
annotator to annotator and from category to category. Table~\ref{tab:CX-purity} shows the CX-purity values for each category and each annotator. The purity numbers varying from 
0.52 to 0.90 for different categories and annotators. 
Different annotators appear to be consistent across categories. 
The same categories, e.g., $j$ and $f$, score lower in purity 
for most annotators.


\begin {table}[h!t]
\centering
\begin{tabular}{|c|c  c c c c |c| } 
 \hline
 Category & \multicolumn{5}{|c|}{Annotator} & \\
 \hline
  & \#1 &\#2 & \#3& \#4 & \#5 &average\\
\hline
a&0.66 & 0.66 & 0.68& 0.65 & 0.65 & 0.66\\
\hline
b&0.69 & 0.76 & 0.75 & 0.76 & 0.78&0.75 \\
\hline
c&0.60 & 0.63 & 0.62 & 0.61 & 0.6 &0.61  \\
\hline
d&0.79 & 0.76 & 0.8 & 0.73 & 0.76&0.77     \\
\hline
e&0.88 & 0.88 & 0.91 & 0.93& 0.90 &    0.90  \\
\hline
f&0.58 & 0.65 & 0.60 & 0.58 & 0.60&   0.60  \\
\hline
g&0.72 & 0.71 & 0.72 & 0.74 & 0.68 &    0.72  \\
\hline
h&0.77 & 0.77 & 0.76 & 0.73 & 0.76 &   0.76     \\
\hline
i&0.62 & 0.61 & 0.61 & 0.52& 0.57 &  0.59     \\
\hline
j&0.56 & 0.61& 0.62 & 0.59 & 0.60 &     0.60   \\
\hline
k&0.80 & 0.75 & 0.71 & 0.75 & 0.76 & 0.75\\
\hline
l&0.87 & 0.88 & 0.85 & 0.86 & 0.86 &0.86\\
\hline
average&  0.66   &0.67  &  0.67  &   0.65   &  0.66  &\\
\hline
\end{tabular}
\caption {CX-purity: Shows the degree to which the activations of a visual concept map to a single X-feature.} \label{tab:CX-purity} 
\end {table}

Table~\ref{tab:XC-purity} shows the XC-purity values for 
different categories and annotators. The purity numbers are 
generally worse, going down to 0.30 in one case (Category $a$, Annotator \#3) and a maximum of 0.83 in a rare case (Category $d$, 
Annotator \#5). This suggests that the 
mapping from X-features to visual concepts 
is more one-to-many than the other way around. 
This is to be expected since the number of 
significant features is often only 2-3 and 
smaller than the number of visual concepts.
Again, different annotators appear to be 
largely consistent across categories with 
the X-features in some categories,
e.g., $d$ and $l$, more easily disentangled than 
the X-features of other categories, e.g., $a$ and $g$. 

\begin {table}[h!t]
\centering
\begin{tabular}{|c|c  c c c c|c|} 
 \hline
 Category & \multicolumn{5}{|c|}{Annotator} &\\
 \hline
  & \#1 &\#2 & \#3& \#4 & \#5& average \\
\hline
a&0.39& 0.51 & 0.30 & 0.45 & 0.38& 0.41\\
\hline
b&0.62& 0.65 & 0.57 & 0.59 & 0.65 &0.62  \\
\hline
c&0.71& 0.55 & 0.49 & 0.79 & 0.51 & 0.61 \\
\hline
d&0.71& 0.8 & 0.75 & 0.77 & 0.84 &0.77 \\
\hline
e&0.71 & 0.7 & 0.66 & 0.73 & 0.65 &0.69 \\
\hline
f&0.72 & 0.44& 0.75 & 0.75 & 0.61 &0.65\\
\hline
g&0.44 & 0.37 & 0.4 & 0.65 & 0.49& 0.46\\
\hline
h&0.46 & 0.46 & 0.5 & 0.49 & 0.38 &0.46\\
\hline
i&0.33& 0.35 & 0.31 & 0.55 & 0.31 &0.37 \\
\hline
j&0.47& 0.56 & 0.58 & 0.54 & 0.55 &0.54\\
\hline
k&0.43 & 0.38 & 0.36 & 0.53 & 0.57 &0.46\\
\hline
l&0.70& 0.81 & 0.61 & 0.75 & 0.86 & 0.75\\
\hline
average&   0.52 &  0.51     &     0.48     &     0.58     &     0.52&\\
\hline
\end{tabular}
\caption {XC-purity: Shows the degree to which the activations of
an X-feature map to a single visual concept.} \label{tab:XC-purity}
\vspace{-1em}
\end {table}

\subsection{RQ3: Inter-annotator Agreement}


 

%
In general we can expect different annotators to produce different namings for a test set, where at least some of the visual concepts differ. Here we consider the extent that these different namings agree and in turn whether explanations produced by different namings are semantically similar. Understanding this issue is important for understanding the extent to which explanations are fundamentally annotator specific. 

\begin{figure}[h!]
\includegraphics[width=8.5cm]{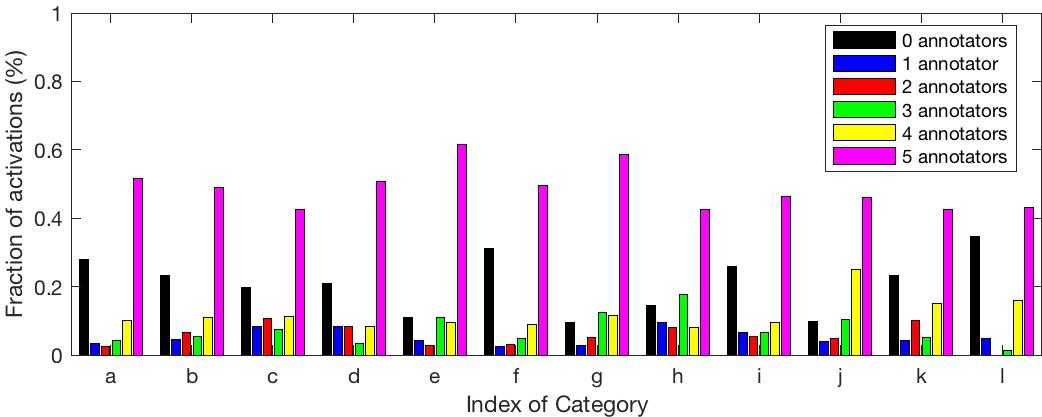}\\
\caption{Fraction of significant activations that are named by exactly $n$ of the annotators, where $n\in \{0, 1,2,3,4,5\}$}
\label{fig:Inter-annotatorAgreement}
\vspace{-1em}
\end{figure}

First, we consider annotator agreement about which significant activations should be named. Figure ~\ref{fig:Inter-annotatorAgreement} shows, for each bird category,
the fraction of significant activations that were named by different numbers of annotators - 0 thru 5. Interestingly, the largest fraction of activations are annotated by all 5 annotators and the second largest are annotated by 0 annotators. This confirms, once again, that for most significant activations, either all annotators choose to assign a name or none of them do. There is strong agreement about the set of activations that should be named.


We now take a more nuanced look at the agreement between 
visual concepts that different annotators created for each category. In particular, we want to know how similar are these concepts and find potential translations between the concepts of different annotators. Are there one-to-one correspondences, are there subsumption relationships, or are there cases of purely incompatible concepts?

Given two namings $N_i$ and $N_j$ produced by two annotators $i$ and $j$, we are interested in matching the clusters, or visual concepts, between the namings. For this purpose, we follow the cluster matching framework of \citep{cazals:hal-01514872} to 
find a best match. The framework first  
defines ``the intersection graph'' $G$ of $N_i$ and $N_j$. $G$ is a bipartite graph where the vertices in each partite set correspond to the clusters of $N_i$ and $N_j$ respectively. There is a weighted edge between each cluster of $N_i$ and $N_j$, with the weight equal to the size of the intersection of the clusters. Thus, a large weight between two visual concepts indicates that they represent many of the same activations in the test set. 



According to \citep{cazals:hal-01514872}, a D-family matching is a partition of all nodes (that belong to either
naming) of the bipartite graph into some number of disjoint sets $S_1, S_2,...$ such that the diameter of all subgraphs of $G$ over the nodes in $S_i$ is $\leq D$. The best D-family matching maximizes the sum of the weights of all edges in all the subgraphs. 


Although restriction of diameter sounds odd, it has a
natural interpretation. 
\textit{D} = 1 corresponds to the standard 1-1 matching between bipartite graphs found by the ``Hungarian algorithm'' \citep{Hungarian-algo}. $D$ = 2 allows one cluster to be refined into multiple clusters on the other side, but disallows further splitting or merging of these smaller sets. Importantly, the clusters can be refined in both directions simultaneously as long as they do not overlap with each other. Thus, when one visual concept in $N_i$ is matched to multiple concepts in $N_j$ for $D=2$, we can view this as an approximate subsumption relationship among the concepts. In this paper we compute agreement scores based on both \textit{D} = 1 and \textit{D} = 2 for each pair of annotators.
\begin{figure*}[h!t]
\centering
\small
\begin{tabular}{c }
  \includegraphics[height=4cm]{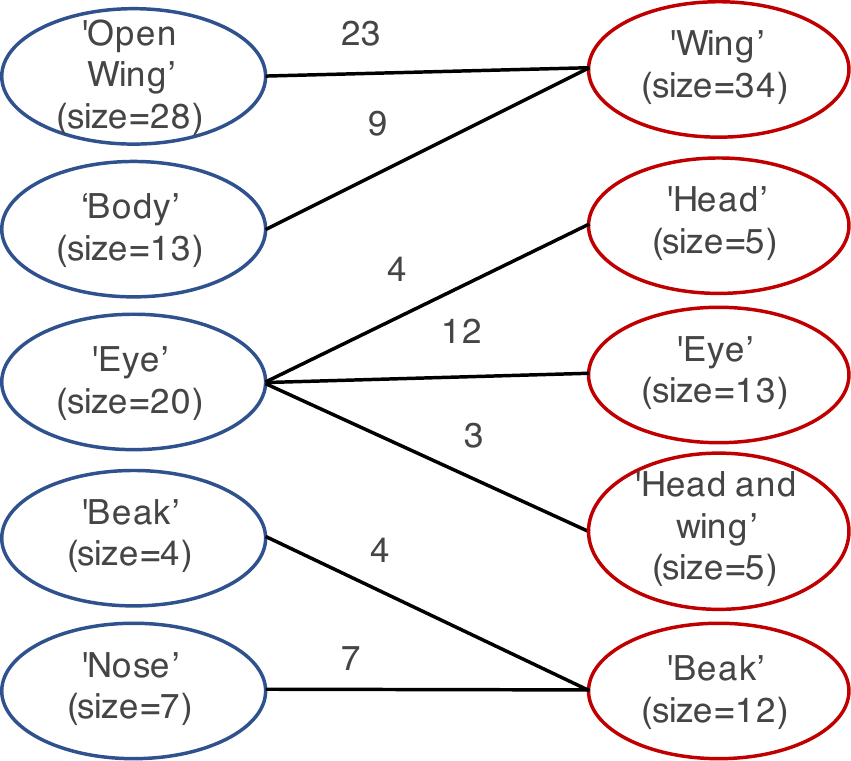}\\
  D = 2\\ \\  \\
  \includegraphics[height=2.5cm]{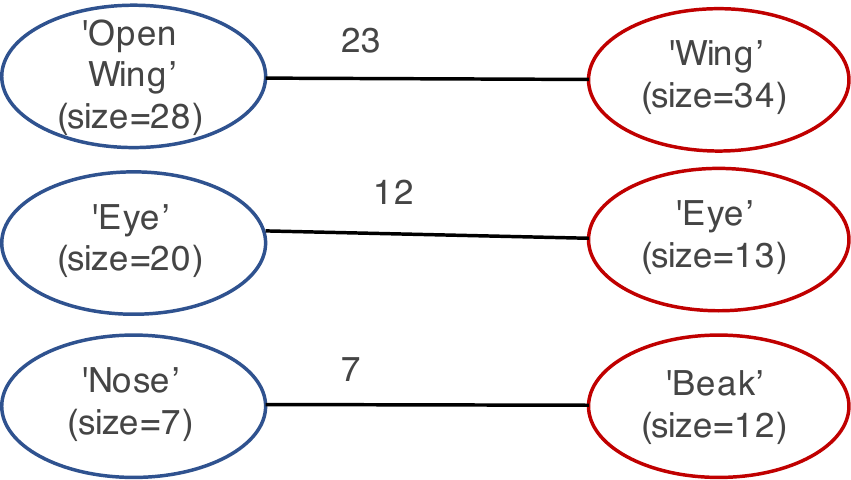}\\
  D = 1\\ \\
\end{tabular}
\begin{tabular}{c }
  \includegraphics[height=4cm]{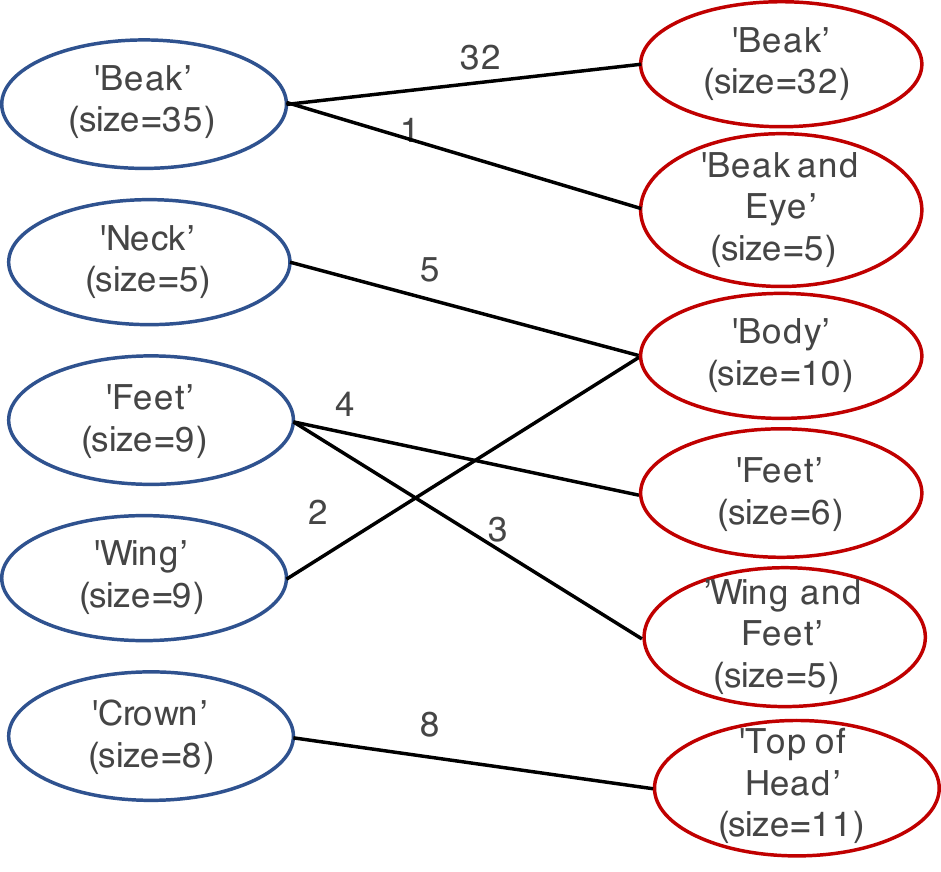}\\
   D = 2\\ \\
  \includegraphics[height=4cm]{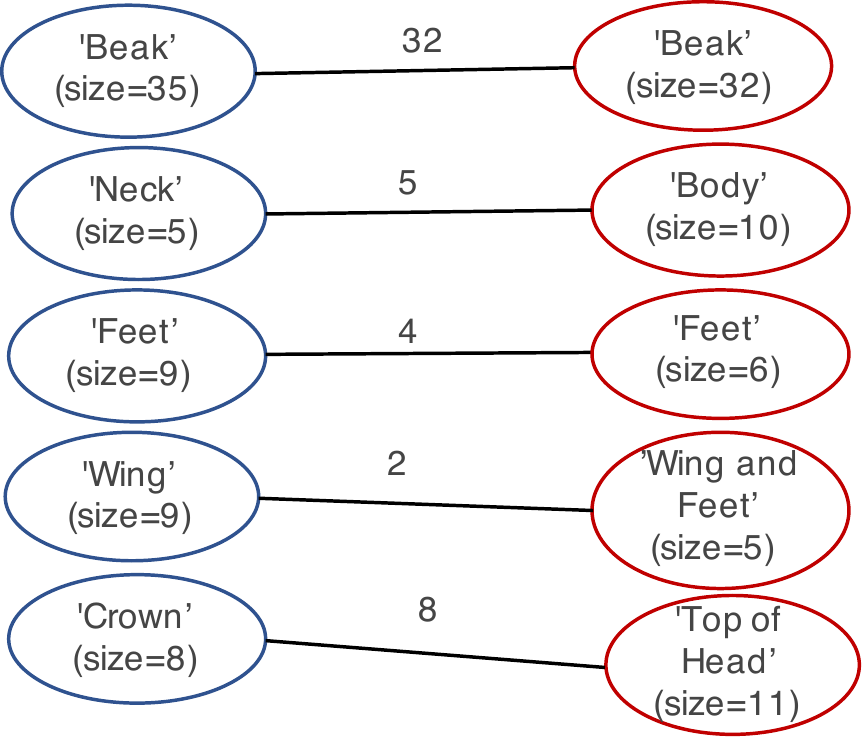}\\
   D = 1\\ \\
\end{tabular}
\begin{tabular}{c}
  \includegraphics[height=4cm]{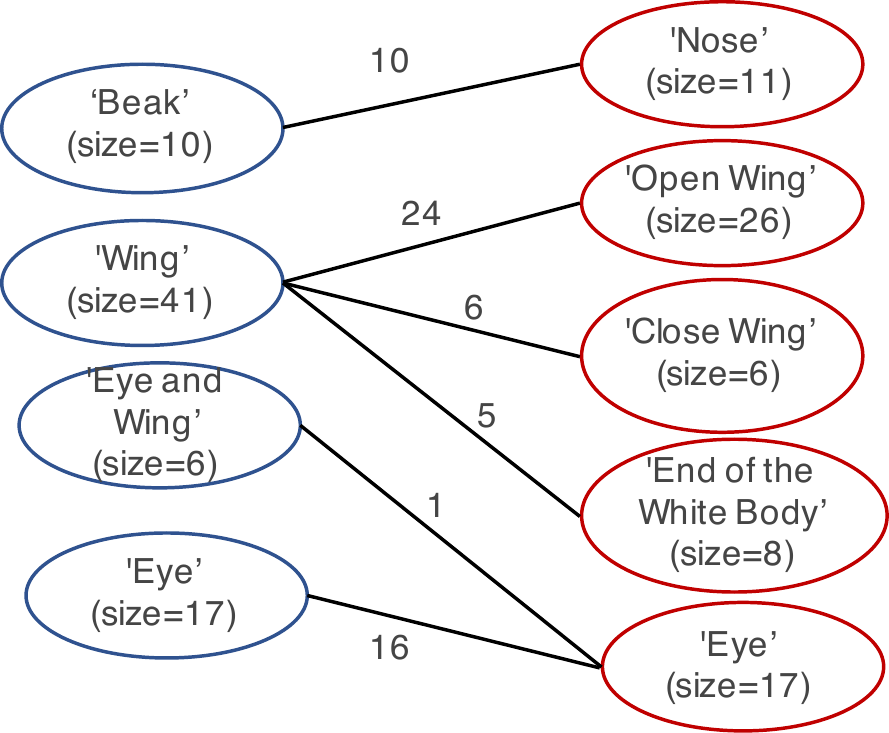}\\
   D = 2\\ \\  \\
  \includegraphics[height=3cm]{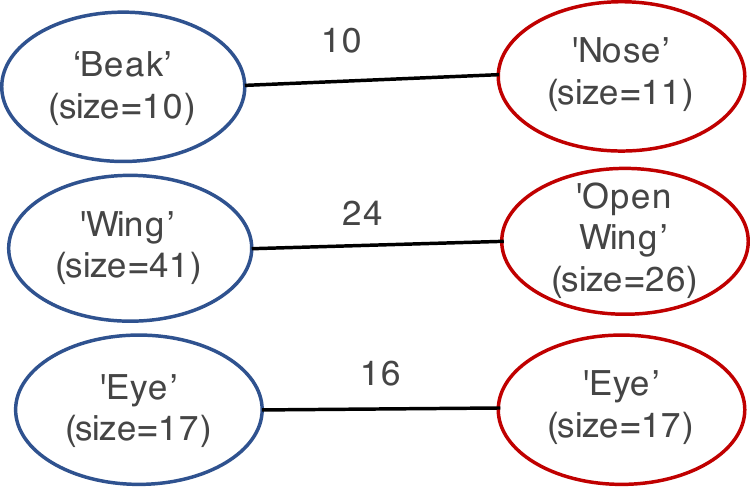}\\
   D = 1\\ \\
\end{tabular}
\caption{Examples of pairwise similarity matching between annotators}
\label{fig:PairwiseGraphs}
\end{figure*}
\begin {table*}[h!t]
\centering
\begin{tabular}{|c|p{15cm} |} 
 \hline
 Category  & Visual concepts combinations\\
 \hline \hline
a  &  ('eye', 'close wing') 56.6667\% , ('close wing') 10\% 
, ('open wing', 'eye') 6.6667\% , ('eye', 'end of the white body') 5\% , ('nose') 5\% , ('eye') 3.3333\% , ('open wing') 1.6667\% , ('end of the white body') 1.6667\% , ('unlabeled')  10\% \\ 
\hline 
b &   ('crown') 25\% , ('beak', 'wing') 25\% , ('wing') 25\% , ('crown', 'neck') 4.5455\% , ('unlabeled') 4.5455\% 
, ('neck') 4.5455\% , ('beak', 'feet') 4.5455\% , ('beak', 'crown') 2.2727\% , ('wing', 'neck') 2.2727\% , ('beak') 2.2727\%  \\ 
\hline 
f  &  ('throat', 'forehead') 64.9123\% , ('throat') 17.5439\% , ('unlabeled') 8.7719\% , ('throat', 'tail') 3.5088\% , ('forehead', 'tail') 3.5088\% , ('forehead') 1.7544\%  \\ 
\hline 

l  &  ('eye') 88.2353\% , ('head') 5.8824\% , ('wing') 1.9608\% , ('unlabeled') 1.9608\% , ('eye', 'wing') 1.9608\%  \\ \hline 
\end{tabular}
\caption {Test set explanation summary for one annotator for a sample of categories} \label{tab:StatisticalAnalysis} 
\end {table*}
Figure~\ref{fig:PairwiseGraphs} shows  examples of matching between 3 pairs of users with \textit{D} = 1 and \textit{D} = 2. It is worth noting that there are one-to-one mappings as well as one-to-many 
mappings from both sides for $D=2$. The words that are associated with each visual concept are the linguistic names assigned by the annotators to those concepts. In the first mapping of Figure 6, for example, both 'Open Wing' and `body' on the left map to `Wing' on the right. On the other hand, the single cluster `Eye' maps to `Head', 'Eye' and 'Head and Wing' on the right. 

Overall, many of the matched visual concepts are sensible based on the linguistic descriptions provided by the annotators. This is especially true for $D=1$. Many of the one-to-many matches in $D=2$ also appear to made sense based on the words. These one-to-many mappings indicate different choices in the resolution of assigning names. There are some other cases, where the matchings do not appear to make semantic sense based on the names. For example, `Eye' maching to `Head and wing' might not appear to be a good match. However, this can simply reflect somewhat arbitrary labeling conventions used by different annotators. For example, one annoator may consider activation maps that are ``brightest'' on the eye to be labeled `eye', while another annotator may also pay attention to parts of the activation map that are not brightest, but still active. 
 
Since not all significant activations are labeled by all annotators, we first try to characterize the 
fraction of common annotations between pairs of  annotators. Let us denote by $A_i$ and $A_j$ the number of significant activations throughout the test set labeled by Annotator \#$i$ and Annotator \#$j$, respectively. we use the Jaccard index, which is the ratio of the
intersection to the union of the two sets, i.e., $\frac {|A_i \cap A_j|} {|A_i \cup A_j|}$, to measure the fraction of the images both annotators 
annotated. This is shown in the last column of  Table~\ref{tab:shortPairwise} averaged over different pairs of annotators. The Jaccard index is fairly high for all categories, indicating that there is a good overlap between the sets of activations chosen by different annotators to annotate.  

We compute the agreement between the two annotators for $D=1$ and $D=2$ as the total weight of all edges in the $D$-family matching as a fraction of the number of activations labeled by both annotators. If we interpret the matchines as translations between namings, then the agreement is the fraction of activations that are translatable between namings. 
The columns labeled ``Agreement'' in 
Table~\ref{tab:shortPairwise} shows the 
statistics of 1-family and 2-family agreements for each category over the set of all
annotator pairs. The average and standard deviation as well as min and max of the scores are shown.  
The agreement numbers are fairly high across most categories, although the minimum values for some categories for $D=1$ are low. 
Since $2$-family matching is more permissive than $1$-family matching, the agreement numbers are higher for $D=2$ as we expect. Even for $D=1$ the agreement in most categories is reasonably high, which shows that there is reason to be optimistic about developing a common ontology for explanations. 




\begin{table}[h!t]
\begin{center}
\begin{tabular}{  |p{1.1cm}|p{1cm}| p{1.5cm} p{1.2cm} p{1.3cm}| } 
\hline
&  &\multicolumn{3}{|c|}{ Pairwise similarity scores } \\
\hline
Category &  & Agreement (D=1) & Agreement (D=2) & Jaccard index\\
\hline
&  min  &0.6&0.88&0.78\\
a & average  &0.74$\pm$0.09&0.96$\pm$0.04&0.82$\pm$0.04\\
&  max  &0.9&1&0.89\\
 \hline 
&  min  &0.74&0.86&0.73\\
b & average  &0.82$\pm$0.05&0.91$\pm$0.04&0.81$\pm$0.04\\
&  max  &0.9&0.97&0.87\\
 \hline 
&  min  &0.7&0.82&0.7\\
c & average  &0.79$\pm$0.07&0.92$\pm$0.05&0.75$\pm$0.05\\
&  max  &0.9&1&0.86\\
 \hline 
&  min  &0.92&0.97&0.75\\
d & average  &0.96$\pm$0.02&0.98$\pm$0.01&0.8$\pm$0.04\\
&  max  &1&1&0.88\\
 \hline 
&  min  &0.79&0.89&0.76\\
e & average  &0.88$\pm$0.06&0.97$\pm$0.04&0.83$\pm$0.05\\
&  max  &1&1&0.91\\
 \hline 
&  min  &0.58&0.82&0.81\\
f & average  &0.77$\pm$0.13&0.94$\pm$0.04&0.86$\pm$0.04\\
&  max  &0.99&0.99&0.91\\
 \hline 
&  min  &0.57&0.77&0.72\\
g & average  &0.69$\pm$0.07&0.91$\pm$0.07&0.8$\pm$0.05\\
&  max  &0.79&1&0.85\\
 \hline 
&  min  &0.59&0.76&0.6\\
h & average  &0.73$\pm$0.06&0.86$\pm$0.07&0.7$\pm$0.08\\
&  max  &0.8&0.95&0.82\\
 \hline 
&  min  &0.6&0.75&0.71\\
i & average  &0.69$\pm$0.07&0.85$\pm$0.07&0.8$\pm$0.04\\
&  max  &0.81&0.94&0.86\\
 \hline 
&  min  &0.57&0.77&0.58\\
j & average  &0.74$\pm$0.1&0.85$\pm$0.05&0.75$\pm$0.13\\
&  max  &0.86&0.92&0.91\\
 \hline 
&  min  &0.52&0.85&0.64\\
k & average  &0.71$\pm$0.1&0.92$\pm$0.05&0.77$\pm$0.08\\
&  max  &0.85&1&0.88\\
 \hline 
&  min  &0.67&0.86&0.74\\
l & average  &0.86$\pm$0.09&0.95$\pm$0.05&0.85$\pm$0.06\\
&  max  &1&1&0.96\\
 \hline 
 \multicolumn{2}{|c|}{ Global Average } &0.78& 0.91 &0.79\\
 \hline
\end{tabular}
\end{center}
\caption {Pairwise comparison between clusters generated by annotators over all categories} \label{tab:shortPairwise} 
\end {table}



{\bf Linguistic Compatibility of Matched Concepts.} We now consider the compatibility between the linguistic labels given to the visual concepts by the annotators
and the semantic mapping found by our matching system. This gives some indication of how well the automated matching computations correspond to the annotator's attempts to attach linguistically meaningful descriptions to concepts.  By convention we will put quotes around individual linguistic names assigned to clusters, which often involve the conjunction of multiple terms. For example, `Wing and Eye', is a name that corresponds to activations that covered the eye and parts of the wing.  

We consider two
types of compatibility -- 1) exact compatibility and  2) subset compatibility -- between 
the linguistic labels given to two matched clusters. 
The word labels are exactly compatible if the words used are considered semantically equivalent: e.g., 'beak' and 'nose', and 'Eye' and 'Eye'. Subset compatibility is when one of the words seems to describe a subset of the other word. For example,  'Open Wing' and 'Wing', or  'Wing' and 'Wing and Eye'. 

The linguistic names that were judged as not compatible can be grouped into two categories: 1) Names that have semantically different meanings, for example, 'Tail' and 'Wing', 'Feet' and 'wing shape' , 'Eye and beak' and 'Body', and 2) Names that are very ambiguous and are difficult to match, for example: 'multiple parts captured' , 'bottom body' ,  'gradinet at feather' ,or  'middle body/Feathers'.

Table~\ref{tab:AgreementNaming} shows the comparison of exact compatibility and subset compatibility  over all categories.
Given a  matching graph of a pair of annotators,  we compute 
the sum of the edge weights for those edges that are semantically compatible divided by the sum of all edge weights.
The numbers in the table are averaged over all pairs of annotators in each category. The Table 
shows that subset compatible word pairs score highly for both $D=1$ and $D=2$ for almost 
all categories. They are also quite consistent across annotators. 
However, exact compatibility between name pairs displays a wide range across annotators 
for almost all categories. This suggests that the meanings of word labels are not exactly consistent across annotators and should not be trusted, although the subsumption relationships 
between the words seem to hold more consistently across the annotators and categories.

\begin{table}[h!t]
\small
\begin{center}
\begin{tabular}{  |p{1cm}|p{0.9cm}| c c |c c| } 
\hline
&  &\multicolumn{2}{|p{1.6cm}|}{Exact compatibility }&\multicolumn{2}{|p{1.6cm}|}{Subset compatibility   } \\
\hline
Category &  &  (D = 1) &  (D = 2)&  (D = 1) &  (D = 2) \\
\hline
&  min  &0.45&0.34&0.9&0.87\\
a & average  &0.65$\pm$0.3&0.56$\pm$0.46&0.99$\pm$0.01&0.95$\pm$0.02\\
&  max  &0.92&0.85&1&1\\
 \hline 
&  min  &0.84&0.75&0.9&0.87\\
b & average  &0.91$\pm$0.03&0.83$\pm$0.04&0.97$\pm$0.01&0.97$\pm$0.02\\
&  max  &1&0.93&1&1\\
 \hline 
&  min  &0&0&0.92&0.9\\
c & average  &0.56$\pm$1.18&0.51$\pm$0.88&0.98$\pm$0.01&0.96$\pm$0.01\\
&  max  &1&0.83&1&1\\
 \hline 
&  min  &0.92&0.87&0.97&0.94\\
d & average  &0.98$\pm$0.01&0.96$\pm$0.01&0.99$\pm$0&0.99$\pm$0\\
&  max  &1&1&1&1\\
 \hline  
&  min  &0.15&0.13&1&1\\
e & average  &0.64$\pm$1.23&0.58$\pm$0.97&1$\pm$0&1$\pm$0\\
&  max  &1&0.92&1&1\\
 \hline 
&  min  &0.19&0.16&1&0.93\\
f & average  &0.37$\pm$0.43&0.35$\pm$0.46&1$\pm$0&0.98$\pm$0.01\\
&  max  &0.93&0.92&1&1\\
 \hline 
&  min  &0.2&0.16&0.92&0.87\\
g & average  &0.61$\pm$0.28&0.49$\pm$0.26&0.99$\pm$0&0.96$\pm$0.01\\
&  max  &0.89&0.8&1&1\\
 \hline 
&  min  &0.09&0.07&0.96&0.97\\
h & average  &0.49$\pm$0.51&0.42$\pm$0.35&0.99$\pm$0&0.99$\pm$0\\
&  max  &0.96&0.86&1&1\\
 \hline 
&  min  &0.06&0.27&0.98&0.97\\
i & average  &0.57$\pm$0.47&0.53$\pm$0.34&1$\pm$0&0.99$\pm$0\\
&  max  &0.97&0.88&1&1\\
 \hline 
&  min  &0.12&0.11&1&1\\
j & average  &0.43$\pm$0.85&0.42$\pm$0.69&1$\pm$0&1$\pm$0\\
&  max  &0.92&0.89&1&1\\
 \hline 
&  min  &0&0&0.37&0.66\\
k & average  &0.3$\pm$0.6&0.32$\pm$0.31&0.94$\pm$0.36&0.95$\pm$0.09\\
&  max  &0.62&0.64&1&1\\
 \hline 
&  min  &0&0&0.59&0.68\\
l & average  &0.35$\pm$1&0.41$\pm$0.77&0.88$\pm$0.31&0.91$\pm$0.18\\
&  max  &1&1&1&1\\
 \hline 
 \multicolumn{2}{|c|}{ Global Average } &0.57 & 0.53 & 0.978 & 0.971\\
 \hline 
\end{tabular}
\end{center}
\caption {Agreement between the automatically produced matchings between clusters (i.e. visual concepts) and the linguistic names assigned to clusters by the annotators.} \label{tab:AgreementNaming} 
\end {table}

{\bf Test Set Explanation Summaries.} One of the motivations for naming a test set is to produce summaries of the explanation types used to predict test images. Table \ref{tab:StatisticalAnalysis} gives examples of test set summaries for 4 categories that are produced based on the namings of a single annotator. For example, we see that for category `a' over 56\% of the test set predictions had the explanation (`eye', `close wing'), which indicates that the network was focusing on the bird eye and closed wing area for those examples. As another example, for category `l' over 88\% of the predictions had the explanation (`eye'), which means the network only looked at the eye area to make the prediction. This type of insight may cause a practitioner to either question the robustness of the classifier if they have reason to believe the eye alone is not discriminative enough. Alternatively, an expert may gain insight from this explanation and realize that the eye is discriminative enough for the task.


\section{Discussion and Conclusions}

In this paper we studied the problem of understanding the decisions of DNNs in terms of 
human-recognizable visual concepts. Our interactive-naming approach involved augmenting the original DNN with a sparser xNN, visualizing the significant activation maps for each decision of the xNN on a test set, and then allowing annotators to flexibly group the activations into recognizable visual concepts, while attaching names to the concepts if desired. The visual concepts can then be used as the basis for producing concise meaningful explanations for test set images. 
We reported on our experience of having 5 annotators use our interface for DNNs trained to recognize different bird species. Our results showed that: 1) annotators were able to assign names to a non-trivial fraction of activations, which allows for at least partial semantic explanations for most test images; 2) the annotators had strong agreement about which activations should and should not be named; 3) there was a non-trivial amount of agreement between the namings produced by different annotators, 
although there were some notable disagreements; and 4) automatically produced matchings, or translations, between different namings had non-trivial agreement with the linguistic names assigned to clusters by the annotators. 


This formative study has set the stage for a variety of future work. Our current interactive naming interface is flexible, but does not attempt to actively reduce the annotator effort. Thus, there is potential to improve the speedup of naming a test set via active learning techniques. We are also interested in interactively training the system based on named concepts, which might reduce the number of activations that cannot be named. In addition, investigations on other datasets with even larger varieties of visual concepts is important for understanding the general characteristics of annotator produced namings.


\bibliography{iui19_InteractiveNaming}
\bibliographystyle{icml2019}

\end{document}